\documentclass[runningheads]{llncs}

\usepackage{eccv}

\usepackage{eccvabbrv}
\usepackage{multirow}
\usepackage{makecell}
\usepackage{marvosym}
\usepackage[marginal]{footmisc}
\usepackage{appendix}
\usepackage{graphicx}
\usepackage{booktabs}

\usepackage[accsupp]{axessibility}  

\usepackage{hyperref}

\usepackage{orcidlink}

\begin{document}

\title{PDT: Uav Target Detection Dataset for Pests and Diseases Tree} 

\titlerunning{PDT Dataset}

\author{Mingle Zhou\inst{1,2,4}\orcidlink{0000-0003-4911-276X} \and
Rui Xing\inst{1,2}\orcidlink{0000-0001-9417-2505} \and
Delong Han\inst{1,2}\orcidlink{0000-0001-7195-3413}\and \\  
Zhiyong Qi\inst{3}\orcidlink{0009-0009-8774-3918} \and 
Gang Li\inst{1,2{(\textrm{\Letter})}}\orcidlink{0000-0002-7896-4833}}

\authorrunning{M. Zhou et al.}

\institute{Key Laboratory of Computing Power Network and Information Security, Ministry of Education, Shandong Computer Science Center (National Supercomputer Center in Jinan), Qilu University of Technology (Shandong Academy of Sciences), Jinan, China\\ 
\email{lig@qlu.edu.cn}\\
\and
Shandong Provincial Key Laboratory of Computer Networks, Shandong Fundamental Research Center for Computer Science, Jinan, China\\
\and
Institute of Remote Sensing and Geographic Information System, School of Earth and Space Sciences, Peking University, Beijing, China\\
\and
SHANDONG SCICOM Information and Economy Research Institute Co., Ltd.
}

\maketitle

\begin{abstract}
  UAVs emerge as the optimal carriers for visual weed identification and integrated pest and disease management in crops. However, the absence of specialized datasets impedes the advancement of model development in this domain.
  To address this, we have developed the Pests and Diseases Tree dataset (PDT dataset). PDT dataset represents the first high-precision UAV-based dataset for targeted detection of tree pests and diseases, which is collected in real-world operational environments and aims to fill the gap in available datasets for this field.
  Moreover, by aggregating public datasets and network data, we further introduced the Common Weed and Crop dataset (CWC dataset) to address the challenge of inadequate classification capabilities of test models within datasets for this field.
  Finally, we propose the YOLO-Dense Pest (YOLO-DP) model for high-precision object detection of weed, pest, and disease crop images.  We re-evaluate the state-of-the-art detection models with our proposed PDT dataset and CWC dataset, showing the completeness of the dataset and the effectiveness of the YOLO-DP.
  The proposed PDT dataset, CWC dataset, and YOLO-DP model are presented at \url{https://github.com/RuiXing123/PDT_CWC_YOLO-DP}. 
\end{abstract}

\section{Introduction}
\label{sec:intro}

The efficacy of intelligent UAV-based plant protection operations hinges on precisely identifying weeds and pests within imagery, representing a critical challenge in computer vision\cite{-2,-1,0}.
As computer vision and UAV technologies have advanced swiftly, the adoption of automated UAV plant protection operations is on the rise\cite{1,2}.
The working mode of plant protection UAV has significant defects: non-intelligent work can result in the misuse or waste of agrochemicals, and the high-intensity, repetitive mechanical tasks can diminish their operational lifespan.
Enhancing the precision of weed and pest target recognition is essential for the effectiveness of intelligent plant protection UAVs.
The absence of specialized datasets and the limitations of existing detection models pose significant constraints on advancing research in this domain.

\begin{figure}[tb]
  \centering
  \includegraphics[width=0.9\linewidth]{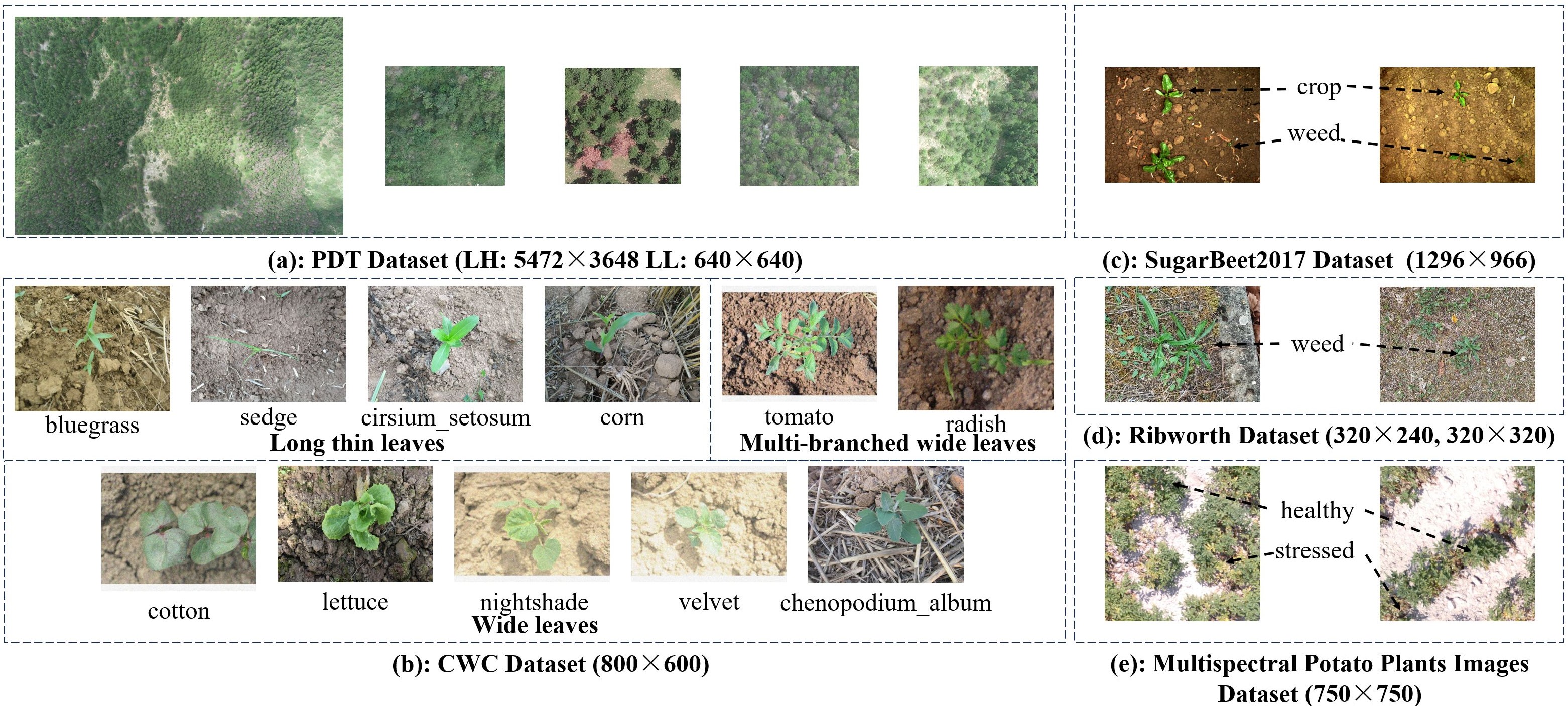}
  \caption{Dataset comparison. 
  (a) shows the PDT dataset (Low Resolution(LL) and High Resolution(LH)): 640×640, 5472×3648.
  (b) shows the characteristics of the CWC dataset: it contains 11 different similar plants.
  (c), (d) and (e) are the public datasets.
  }
  \label{fig:shujvji}
\end{figure}

Drones typically take images at both high and low altitudes when outdoors.
Existing datasets are usually taken from indoor greenhouses (\cref{fig:shujvji} (c)), where the majority of the data consists of target images captured at close range (\cref{fig:shujvji} (d)). 
Such data struggles to incorporate environmental factors like changes in lighting, and the target sizes significantly deviate from real-world conditions, failing to satisfy practical requirements.
Further, the majority of existing datasets are limited to single or double classes, which cannot meet the training of model classification ability, as illustrated in \cref{fig:shujvji} (c), (d), and (e).

On the other hand, the conventional detection models utilized in UAV-based intelligent plant protection could be more challenging in accurately identifying targets.
The deficiency in this field is underscored by the absence of tailored baseline models, resulting in unmet demands for detecting and managing diverse weeds, crops, pests, and diseases.
This paper aims to address the lack of dedicated datasets and algorithms for weeds, pests and diseases in agricultural drones.
Overall, the main contributions of this paper are as follows:
\begin{itemize}
\item We have developed the UAV target detection agricultural Pests and Diseases Tree dataset, known as PDT dataset.
\item We collated public data and provided Common Weed and Crop dataset (CWC dataset) to train the classification capabilities of the model. 
\item We have designed a specialized detection model, YOLO-DP, tailored for the dense object detection of tree crop pests and diseases. 
\item We reevaluated and analyzed existing generic and specific detection models on PDT and CWC datasets to establish benchmarks for the field.
\end{itemize}

\section{Related Work}

\textbf{Agricultural UAV Detection Dataset.}
In the realm of agricultural robotics, acquiring pertinent datasets frequently presents a challenge, primarily due to the necessity of maintaining dedicated experimental fields and the critical timing of data collection, which is pivotal to the research outcomes.
Chebrolu et al. presented SugarBeet2016 large agricultural robot dataset and subsequently introduced the SugarBeet2017 dataset tailored for the detection task\cite{3,4}.
The limitation of this dataset is that the data, captured from the vantage point of a ground robot, significantly diverges from the actual detection scenarios encountered by UAVs. 
Vakanski et al. introduced a dataset of multispectral potato plant images taken by drones\cite{5}. 
The limitation of this dataset is the lack of raw data, consisting of only 360 images and a total of 1,500 images produced by data enhancement, which cannot effectively guide the model to learn the details of the crop.
The University of Burgundy has launched the Ribworth Dataset, designed for crops and weeds, which uses data augmentation to simulate real-world environmental conditions\cite{6}. 
The limitation of this dataset is that its acquisition target is relatively close, and the factors such as light change are not included, which is not enough to meet the requirements of accurate detection tasks of agricultural plant protection UAVs.

To foster the advancement of precision agriculture, we introduce the PDT dataset and CWC dataset, aimed at addressing the requirements for detecting agricultural weeds, crops, and pests.

\textbf{Agricultural UAV Detection Model.}
The agricultural UAV target detection task has garnered significant attention within the realms of machine learning and computer vision, owing to its attributes of precise localization, efficient operation, and minimal environmental impact.
Among them, the weed detection models based on YOLO deep learning model include: Zhang et al. developed a detection model for weed location and identification in wheat fields by using YOLOv3-tiny\cite{7}. 
Gao et al. modified the YOLOv3-tiny model and applied it to weed detection in sugar beet fields. 
Jabir and Falih proposed an intelligent wheat field weed detection system based on YOLOv5\cite{9}.
Furthermore, a variety of neural network architectures have also been recognized by researchers and implemented in the field of precision agriculture.
Guo et al. proposed WeedNet-R based on the RetinaNet architecture for weed identification and location in sugar beet fields\cite{4}.
Tang et al. employed Kmeans feature learning in conjunction with a convolutional neural network for the recognition of weeds\cite{10}.
Agarwal et al. utilized the Kmeans clustering algorithm to analyze color space features in multispectral weed images for effective weed detection\cite{11}.

It is evident that there is a scarcity of studies focusing on baseline models for the detection of tree diseases and pests, particularly those related to the invasion of the exotic species Red Turpentine Beetle.
Furthermore, the majority of these baseline models are trained using single-class or two-class detection datasets, and the classification ability of the models is limited by the diversity of the data.
However, for the detection tasks in large-scale, high-precision agricultural UAV operations, precise detection, accurate classification, and the expansion of crop detection varieties are essential. 
Consequently, we propose the YOLO-DP detection model to address these requirements.

\section{PDT Dataset}

In this study, we introduced a PDT dataset with both high and low resolution versions, specifically for the detection of red turpentine beetle natural pest targets.
As most existing public datasets are gathered indoors or via simulated drones, they fail to capture the characteristics of outdoor UAV detection environments.
Consequently, there is an immediate demand for an agricultural dataset that is tailored to real-world operational environments and encompasses the distribution of both high and low-altitude data, thereby fulfilling the high-precision, large-scale target detection requirements from a UAV perspective.
This section delineates the data acquisition process for the PDT dataset, encompassing the selection of the data collection domain, the equipment used for acquisition, and the criteria for sample definition.

\subsection{Data Acquisition}

\textbf{Selection of the Data Collection Domain.}
The Red Turpentine Beetle, a member of the Coleopteridae family, not only invades extensive pine forest areas but also contributes to the widespread mortality of trees, exerting a profound impact on the regional ecological environment and forestry.
Despite the implementation of preventive and control strategies, including monitoring and trapping, effectively managing its continued proliferation remains a formidable challenge.
The precision spraying platform of plant protection UAV is an effective solution, but the public UAV data for the invasive species Red Turpentine Beetle is blank.
Consequently, we opted to collect data from a cluster of dead pine trees that have been infested by the Red Turpentine Beetle.

\textbf{High-resolution Drone Camera Equipment.}
To obtain accurate sample data, we use a UAV mapping camera called DJI-ChanSi L2.
This advanced system enhances the UAV flight platform's capabilities for acquiring accurate, efficient, and reliable 2D and 3D data. 
The LiDAR components have excellent performance, ranging accuracy of 2 cm at a height of 150 meters, laser wavelength of 905 nm, and laser pulse emission frequency of 240 kHz.
The system supports the exFAT file system and can capture images in JPEG or PNG formats.
In essence, the DJI-ChanSi L2 is a comprehensive platform that integrates lidar, a high-precision navigation system, and a mapping camera, specifically designed for high-precision data acquisition in mapping, forestry, power grid, and engineering and construction industries.

\begin{figure}[tb]
  \centering
  \includegraphics[width=0.75\linewidth]{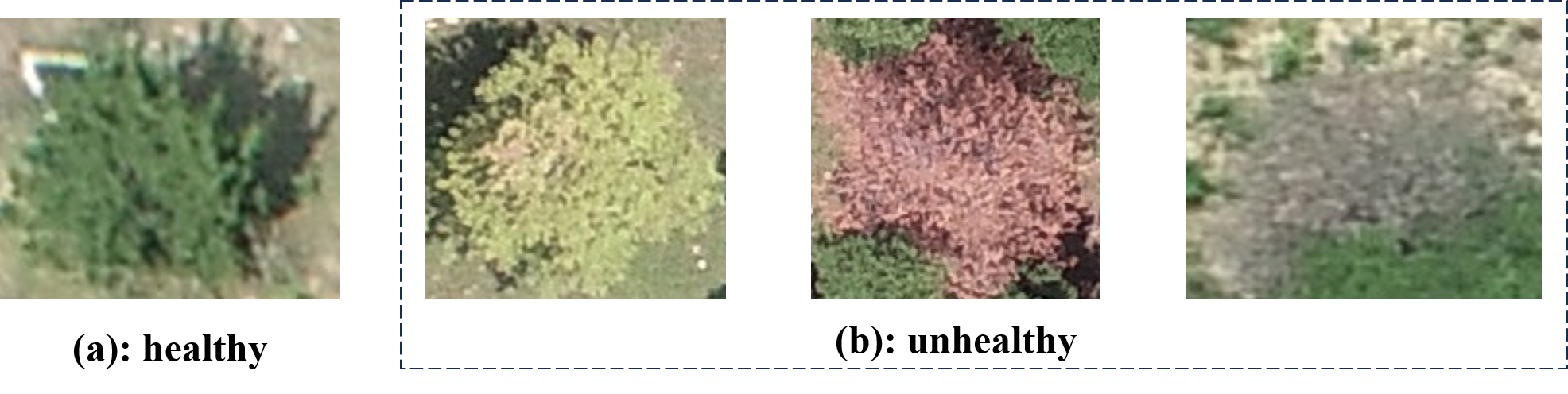}
  \caption{Data example.
  (a) is a healthy goal and (b) is a unhealthy goal. 
  The PDT dataset takes (b) as the category.}
  \label{fig:mubiao}
\end{figure}

\textbf{Definition of Detection Target.}
Distinct from close-range pest detection, the task of pest detection at low or high altitudes fundamentally involves capturing the differential characteristics between the affected vector and the healthy state.
Consequently, the detection samples for the PDT dataset are sourced from both healthy and unhealthy trees within pine forests.
As depicted in \cref{fig:mubiao}, (a) illustrates a pine tree in a healthy state, while (b) displays three pine trees in a unhealthy state that have been infested by pests, exhibiting progressively severe symptoms from left to right.
\begin{figure}[tb]
  \centering
  \includegraphics[width=0.9\linewidth]{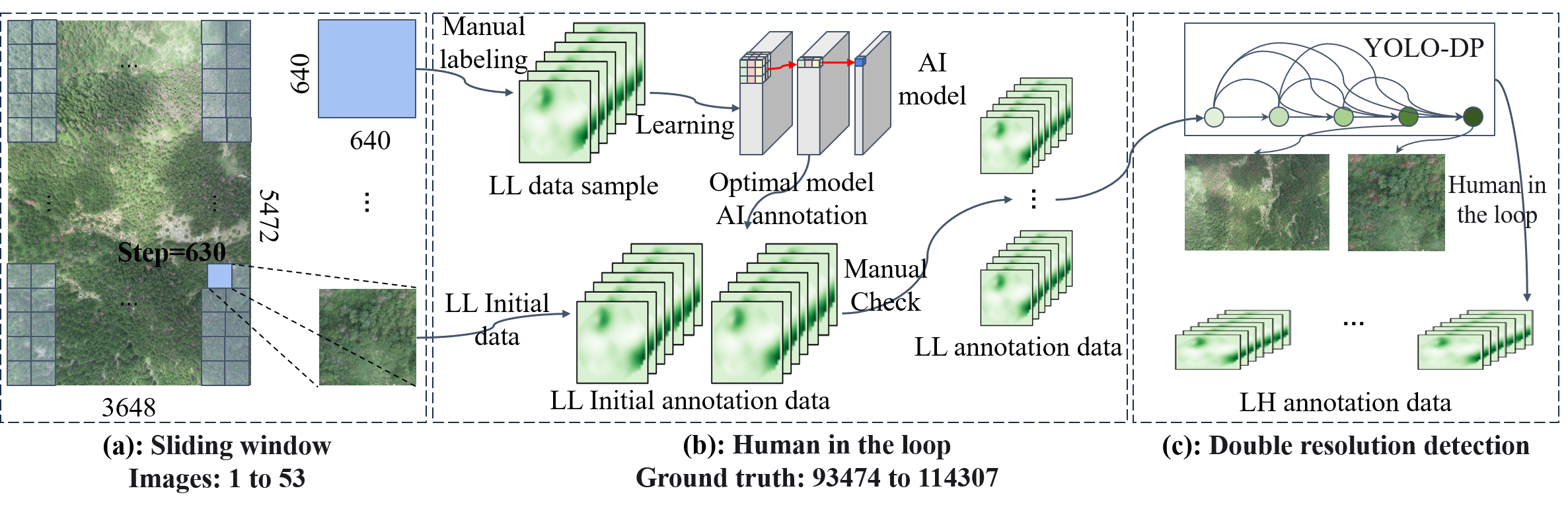}
  \caption{PDT dataset generation and detection process. 
  (a) represents the sliding window method, and (b) represents the ``Human-in-the-loop'' data annotation method.
  (c) means that a LL image is sent to the neural network for training, and LL and LH dual-resolution images are detected at the same time.}
  \label{fig:jianceliucheng}
\end{figure}

\subsection{Data Preprocessing and Statistics}
This section delves into the acquisition, processing, and production procedures of the original image data about the PDT dataset.
We set up an isometric photo mode at an altitude of 150 meters to acquire 105 sets of raw image data and 3D point cloud data.
Owing to the substantial size of the original images, they are unsuitable for training with the current mainstream object detection neural networks.
We introduce a training approach that applies to the majority of neural network data.
As illustrated in \cref{fig:jianceliucheng}, we employ a methodology that encompasses ``data preprocessing - `Human-in-the-loop' data labeling - manual correction''.
The essence of this method is to leverage the preliminary analysis capabilities of artificial intelligence to aid human annotators, thereby enhancing the efficiency and accuracy of data annotation.

\textbf{Data Preprocessing.}
\cref{fig:jianceliucheng} (a) demonstrates the application of the sliding window technique on the original image data, which allows for the extraction of a usable, standard-sized image through regional cropping of the high-resolution image.
To prevent information loss during the sliding window process, the window dimensions are set to 640×640 pixels, with a step size of 630×630 pixels.
Following the sliding window operation, 53 LL images were derived from a single LH image.
Furthermore, due to the high similarity between the unhealthy pine tree depicted in \cref{fig:mubiao} and the ground background, and to enhance the effectiveness of neural network training, we opt to retain images that do not contain targets.
Upon generating an output from these data during training, a corresponding loss is produced, which aids the neural network in learning and reduces the rate of false positives. 
\begin{table}[tb]
  \caption{Structure of PDT dataset and its sample.}
  \begin{center}
  \label{tab:1}
  \resizebox{0.8\linewidth}{!}{
  \begin{tabular}{@{}cccccccccc@{}}
    \toprule
    \multirow{2}{*}{Edition} & \multirow{2}{*}{Classes} & \multirow{2}{*}{Structure} & \multirow{2}{*}{\makecell{Targeted\\images}} & \multirow{2}{*}{\makecell{Untargeted\\images}} & \multirow{2}{*}{\makecell{Image\\size}} & \multirow{2}{*}{Instances} & \multicolumn{3}{c}{Target Amount}\\
    & &  &  &  &  &  & S(Small) & M(Medium) & L(Large)\\
    \midrule
    \multirow{2}{*}{Sample} & \multirow{2}{*}{unhealthy}  & Train & 81 & \multirow{2}{*}{1} & \multirow{2}{*}{640$\times$640} & 2569 & 1896 & 548 & 179\\
     & & Val & 19 &  & & 691 & 528 & 138 & 25\\
     \hline
     \multirow{3}{*}{LL} &  \multirow{3}{*}{unhealthy} & Train & 3166 & 1370 & \multirow{3}{*}{640$\times$640} & 90290 & 70418 & 16342 & 3530\\
     &  & Val & 395 & 172 & & 12523 & 9926 & 2165 & 432\\
     &  & Test & 390 & 177 &  & 11494 & 8949 & 2095 & 450\\
     \hline
    LH & unhealthy & - & 105 & 0 & 5472$\times$3648 & 93474 & 93474 & 0 & 0\\
  \bottomrule
  \end{tabular}}
  \end{center}
\end{table}

\textbf{Data Annotation.}
We proceed to annotate the preprocessed data. To ensure the data's validity, we utilize the widely-used Labelme \footnote[1]{https://github.com/labelmeai/labelme} data labeling software.
For high-resolution datasets with dense, small targets, the sheer volume of targets per image makes manual annotation a tedious and time-intensive process.
As illustrated in \cref{fig:jianceliucheng} (b), the ``Human-in-the-loop'' data annotation approach effectively addresses this challenge.
Initially, we acquire the LL version samples of the PDT dataset through manual annotation.
The structure of the sample dataset is depicted in \cref{tab:1}, which comprises a training set and a validation set.
Subsequently, the sample dataset is input into the YOLOv5s model for training over 300 epochs, with the optimal weights being saved.
To ensure that no label data is missed, a confidence threshold of 0.25 and an NMS IOU (Intersection over Union) threshold of 0.45 were set to automatically label the LL data and obtain the original label set.

\textbf{Manual Check.}
The labeled dataset underwent manual filtering to eliminate incorrect annotations, rectify erroneous ones, and add any omitted annotations.
The LL datasets were generated in two widely recognized formats, YOLO\_txt and VOC\_xml, by automatically partitioning the data structure into an 8:1:1 ratio using scripts.
For the LH data, we input the completed LL data into the YOLO-DP model for training to acquire the optimal weights.
The ``Human-in-the-loop'' data annotation and subsequent manual verification were conducted on the LH data to obtain the final LH version dataset. 

\textbf{Data Statistics.}
The statistics of the PDT dataset are shown in \cref{tab:1}.
LL and LH versions consist of 5,670 and 105 images, respectively, with 114,307 and 93,474 samples, and feature a single class labeled as unhealthy.
The training, validation, and test sets of the LL version include 1,370, 172, and 144 target-free images, respectively.
On average, the LL and LH versions contain 29 and 890 samples per image, respectively, indicating that the PDT dataset is suitable for the task of high-density and high-precision inspection of plant protection drones.

\section{CWC Dataset}
We compile the available data to construct the Common Weed and Crop datasets.
In special cases, there may be crop crossovers (usually in small trial plots) where weeds and crop targets exhibit multi-class structural characteristics and similarities.
The majority of existing public datasets are limited to single or double classes, which do not suffice for the training and classification capabilities required by models.
Consequently, when training the neural network, it is essential to supply detailed texture data that differentiates between various objects, thereby equipping the model with robust classification capabilities. 
This enables the model to effectively sift through and eliminate highly similar negative samples during detection, thereby enhancing the model's detection accuracy.
This section discusses the acquisition and processing of CWC original image data.

\subsection{Data Sources}
In \cref{tab:3}, we present the source, categories, initial quantities of the CWC raw data, as well as their respective resolutions.
As depicted in \cref{fig:shujvji} (b), the CWC original data comprises a total of 11 categories, characterized by similar size, texture, and color.
Please note that we manually annotated the raw Corn weed datasets, Lettuce weed datasets, and Radish weed datasets, while the Fresh-weed-data was already pre-labeled and available in YOLO\_txt format.
\begin{table}[tb]
  \caption{CWC dataset sources.}
  \begin{center}
  \label{tab:3}
  \resizebox{0.8\linewidth}{!}{
  \begin{tabular}{@{}cccccccc@{}}
    \toprule
    Datasets & \makecell{Corn weed\\datastes\cite{LAR}} & \makecell{lettuce weed\\datasets\cite{LAR}} & \makecell{radish weed\\datasets\cite{LAR}} & \multicolumn{4}{c}{Fresh-weed-data\cite{12}}\\
    \midrule
    Classes & \makecell{bluegrass, corn, sedge, \\chenopodium\_album, \\cirsium\_setosum} & lettuce & radish &  nightshade & tomato & cotton & velvet\\
    Number & 250 & 200 & 201 & 115 & 116 & 24 & 38\\
    Image Size & 800$\times$600 & 800$\times$600 & 800$\times$600 & 800$\times$600 & 800$\times$600 & 586$\times$444 & 643$\times$500\\
    \bottomrule
  \end{tabular}}
  \end{center}
\end{table}

\subsection{Data Preprocessing and Statistics}
\textbf{Data Preprocessing.}
Owing to the imbalance between positive and negative samples in the original CWC data, there is a risk that the model may become biased towards the majority class, leading to a diminished recognition capability for the minority class.
We used the oversampling strategy, using the data augmentation method to equalize the number of samples.
The methods include random rotation, random translation, random brightness change, random noise addition, flipping and cutout.
This enhances the robustness of the dataset, simulating the illumination formula: $\boldsymbol{I }_{out}=\boldsymbol{I}_{in}\times w + [\boldsymbol{255}] \times (1-w)$,  $w\in \left[0.35,1 \right]$.
Where $\boldsymbol{I}_{out}$, $\boldsymbol{I}_{in}$ is the output and input images. $w$ is the random weight.

\textbf{Data Statistics.}
The statistics for the CWC dataset are presented in \cref{tab:4}.
The CWC dataset comprises 2,750 images and 2,599 data entries, with the majority featuring large-sized objects.
Data enhancement makes the sample reach a balanced distribution, with an average image containing 1 to 2 objects.
\begin{table}[tb]
  \caption{CWC dataset structure.}
  \begin{center}
  \label{tab:4}
  \resizebox{0.9\linewidth}{!}{
  \begin{tabular}{@{}ccccccccccccc@{}}
    \toprule
    Classes &  & bluegrass & \makecell{chenopodium\_\\album} & \makecell{cirsium\_\\setosum} & corn & sedge & lettuce & radish & nightshade & tomato & cotton & velvet\\
    \midrule
    \multirow{3}{*}{\makecell{Targeted\\Images}} & Train & 200 & 200 & 200 & 200 & 200 & 200 & 200 & 200 & 200 & 200 & 200\\
    & Val & 40 & 40 & 40 & 40 & 40 & 40 & 40 & 40 & 40 & 40 & 40\\
    & Test & 10 & 10 & 10 & 10 & 10 & 10 & 10 & 10 & 10 & 10 & 10\\
    \hline
    \multirow{3}{*}{\makecell{Targeted\\Amount}} & S & 1 & 0 & 0 & 5 & 0 & 0 & 0 & 0 & 0 & 0 & 3\\
    & M & 0 & 0 & 0 & 9 & 0 & 0 & 0 & 0 & 0 & 0 & 0\\
    & L & 249 & 250 & 250 & 236 & 250 & 444 & 326 & 250 & 210 & 268 & 248\\
    \hline
    \makecell{Image\\Size} &  & 800$\times$600 & 800$\times$600 & 800$\times$600 & 800$\times$600 & 800$\times$600 & 800$\times$600 & 800$\times$600 & 800$\times$600 & 800$\times$600 & 586$\times$444 & 643$\times$500\\
    \bottomrule
  \end{tabular}}
  \end{center}
\end{table}

\section{Dataset Comparison}
\cref{tab:5} provides a thorough comparison of the datasets.  
The LH version of the PDT dataset is exceptionally clear, offering image quality that is 35 to 200 times superior to other publicly available datasets.
The PDT dataset is characterized by density, small target and rich real environmental details, which accords with the typical environmental factors of high and low altitude UAV work, and is suitable for training the special UAV detection model for plant protection.
Furthermore, the PDT dataset includes a 3D point cloud version. While not highlighted in this study, a 3D data version is planned for development and will be released to the public in the future.

On the other hand, the CWC dataset demonstrates outstanding classification performance, particularly in terms of category diversity, detailed texture information, and dataset authenticity. 
CWC dataset has a large number of categories, 2 to 5 times more than other publicly available datasets, which indicates that the CWC dataset is well suited for tasks involving high-precision classification of plant protection UAVs.
\begin{table}[tb]
  \caption{Dataset comparison.}
  \begin{center}
  \label{tab:5}
  \resizebox{0.9\linewidth}{!}{
  \begin{tabular}{@{}cccccccccccc@{}}
    \toprule
    Dataset & Resolution & Classes & \makecell{Rich \\Details} & \makecell{High \\Definition} & \makecell{Scale} & \makecell{Dense \\Target} & \makecell{Target \\Scale} & \makecell{Uav \\Collection} & \makecell{No Target \\Image} & \makecell{3D Point \\Cloud Data} & \makecell{Annotation \\Quality}\\
    \midrule
    \multirow{2}{*}{PDT dataset} & 5472$\times$3648 & 1 & - & \checkmark & 100$+$ & \checkmark & S & \checkmark & \checkmark & \checkmark & \multirow{2}{*}{\makecell{``Human-in-\\the-loop''}}\\
     & 640$\times$640 & 1 & - & - & 5K$+$ & \checkmark & S/M/L & \checkmark & \checkmark & \checkmark & \\
     CWC dataset & 800$\times$600 & 11 & \checkmark & - & 2K$+$ & - & L & - & - & - & Manually\\
     SugarBeet2017\cite{4} & 1296$\times$966 & 2 & - & - & 5K$+$ & - & S/M & - & - & \checkmark & Manually\\
     \makecell{Multispectral\\Potato Plants\\Images (MPP)\cite{5}} & 750$\times$750 & 2 & - & - & 1K$+$ & \checkmark & L & \checkmark & - & - & Manually\\
     \makecell{Ribworth\\Dataset (RI)\cite{6}} & \makecell{320$\times$240\\320$\times$320} & 1 & \checkmark & - & 5K$+$ & - & M/L & - & - & - & Manually\\
     \makecell{Corn weed\\Dataset\cite{LAR}} & 800$\times$600 & 5 & \checkmark & - & 6K$+$ & - & L & - & - & - & No\\
     \makecell{lettuce weed\\Dataset\cite{LAR}} & 800$\times$600 & 2 & \checkmark & - & 700$+$ & - & L & - & - & - & No\\
     \makecell{radish weed\\Dataset\cite{LAR}} & 800$\times$600 & 2 & \checkmark & - & 400$+$ & - & L & - & - & - & No\\
     \makecell{Fresh\\-weed-data\cite{10}} & \makecell{800$\times$600\\586$\times$444\\643$\times$500} & 4 & \checkmark & - & 200$+$ & - & L & - & - & - & Manually\\
     \makecell{crop and weed\\detection data\\(CAW)\cite{CAW}} & 512$\times$512 & 2 & \checkmark & - & 1K$+$ & - & L & - & - & - & Manually\\
     Weeds\cite{weeds} & 480$\times$480 & 1 & - & - & 600$+$ & - & S/M & \checkmark & - & - & Manually\\
    \bottomrule
  \end{tabular}}
  \end{center}
\end{table}

\begin{figure}[tb]
  \centering
  \includegraphics[width=0.9\linewidth]{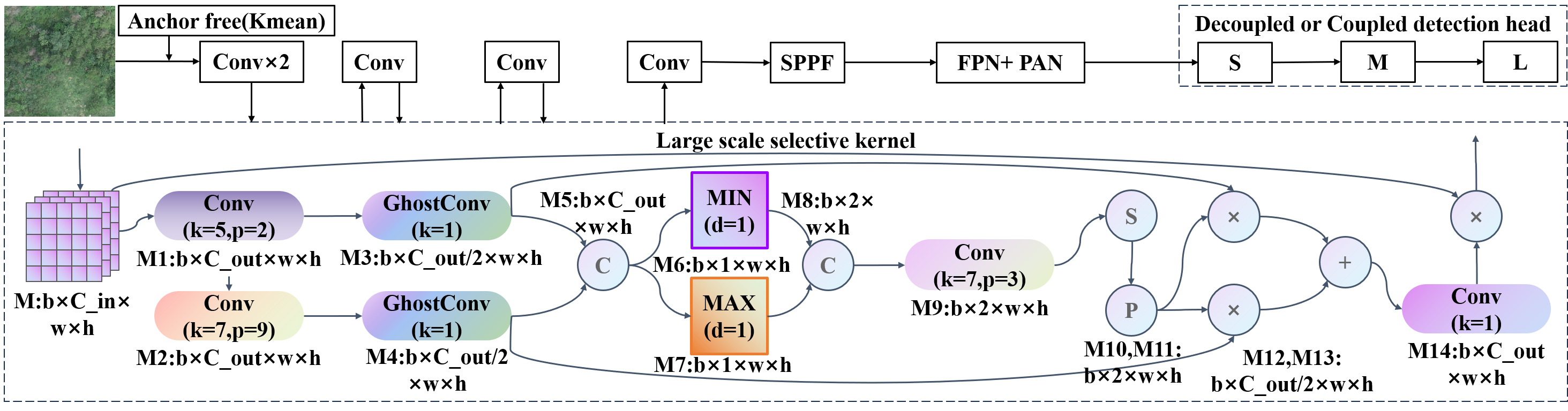}
  \caption{YOLO-DP baseline model architecture. 
  The FPN\cite{FPN}+PAN\cite{PAN} module consists of GhostConv\cite{GhostConv}, Upsample, Concat, and C3. 
  C stands for Concat, S for Sigmoid, P for channel number dilation, $\times$ for matrix multiplication, and $+$ for matrix addition.}
  \label{fig:jiagou}
\end{figure}

\section{YOLO-DP Model}
To expedite research efforts in UAV detection models for intelligent agricultural plant protection, we have designed and proposed a model, YOLO-DP (\cref{fig:jiagou}), specifically dedicated to the detection of tree crop diseases and pests, building upon the YOLOv5s foundatin\cite{YOLOv5}.
To ensure the scalability of the model, the Kmeans clustering algorithm is employed to determine the distribution of the actual bounding box data, allowing for the dynamic adjustment of the anchor box sizes.
In light of the UAV's large-scale detection capabilities, an adaptive Large Scale Selective Kernel is incorporated into the Backbone network to capture the location information of dense, small target pest-infested trees\cite{13}.
Taking into account the efficient detection mode of UAVs, GhostConv is utilized in the Neck network to reduce the model size and computational complexity. Concurrently, the receptive field is enlarged to capture more texture feature information, thereby enhancing the model's classification capabilities.
A version with decoupled detection heads is provided to minimize the interference between classification and regression losses, allowing the model to focus on specific tasks, improving the accuracy and accuracy of the detection, and also improving the model's generalization ability.

\textbf{Adaptive Large Scale Selective Kernel.}
To tackle the demanding detection task involving a broad range of dense, small targets, we gather pest and disease target information by designing a Convolutional group with an extensive receptive field.
Firstly, the matrix M obtains the shallow range information M1 by Conv with $Kernel size (k)=5$ and $Padding (p)=2$.  
The depth information is explored for M1, and the deep range information M2 is obtained using Conv with $k=7$ and $p=9$.  
Feature selection was performed on shallow and depth range information, and M3 and M4 were obtained by grouping convolution on M1 and M2 using GhostConv with $k=1$.
The advantage of using GhostConv is that half of the range information is retained and the other half is processed to prevent excessive loss of information.
Next, the spatial attention is calculated for M3 and M4 large-scale information.
M3 and M4 are concatenated in the channel dimension to obtain M5, and M6 and M7 are obtained by averaging and maximizing the channel dimensions, respectively.
M6 and M7 are concatenated in the channel dimension to obtain M8 with scale $b\times1\times w\times h$, which is fed into Conv with $k=7$ and $p=4$ for range attention collection to obtain M9.  
After Sigmoid activation, the channel dimension is expanded to obtain M10 and M11.
Matrix multiplication is performed with M3 and M4 to obtain M12 and M13, respectively.
After bitwise addition, the final spatial attention matrix M14 is obtained after Conv ($k=1$).  
Finally, the output is obtained by matrix multiplication with the input matrix.

\section{Experiment}
\subsection{Experimental Conditions}
The experimental environment of this paper is ubuntu 18.04, Python3.9, PyTorch 1.9.1 framework, CUDA 11.4 version, and cuDNN 8.0 version.
This paper’s model training and inference are performed on NVIDIA A100-SXM4 with 40GB GPU memory and 16 GB CPU memory on the experimental platform.
The training metricss in this paper are: 300 epochs of iterative training, the early stopping mechanism is set to 100 rounds, the initial learning rate is 0.01, the weight decay of 0.001 and momentum of 0.9\% are used, the Batch-size is 16. 

Experiments were carried out on the PDT dataset and CWC dataset presented in this paper, as well as on the public dataset SugarBeet2017. 
We test the performance of the model YOLO-DP according to the characteristics of the dataset. 
First, the PDT dataset is used to test the detection ability of the model for dense small targets in the UAV perspective. 
Secondly, the ability of the model to extract fine-grained texture information is measured on the CWC dataset to verify its classification ability. It is reflected in the P metrics.
Finally, the detection accuracy was measured on the SugarBeet2017 dataset. It is reflected in the R metrics.
In order not to lose the image information, the input sizes are 640×640, 800×800, 1296×1296 respectively.

\subsection{Experimental Analysis}
\textbf{Dataset Validation.}
We evaluated 6 detection models on seven datasets (2 of ours and 5 public datasets mentioned in the paper).  
Based on the dataset's characteristics, we choose different metrics for the ranking model (Rank (Metrics)).  
Specifically, dense target datasets collected in natural environments use the comprehensive F1 metrics, datasets with target size changes use the R metrics, and multiclass datasets use the P metrics.  
We sort models with the same metrics score again using Gflops.  
Our model has excellent performance on all 7 datasets, and the performance of the proposed 2 datasets differs from that of existing datasets, indicating that our dataset can provide unique data distribution to the research field (\cref{tab:5.0}).
\begin{table}[ph]
  \caption{Datasets effect ranking.}
  \label{tab:5.0}
  \centering
  \resizebox{0.7\linewidth}{!}{
  \begin{tabular}{@{}cc|cc|cc|ccc@{}}
    \hline
     \multirow{2}{*}{Method} & \multirow{2}{*}{Gflops} & \multicolumn{2}{c|}{ Rank (F1)} & \multicolumn{2}{c|}{ Rank (P)} & \multicolumn{3}{c}{ Rank (R)}\\
      &  & PDT (LL) (our) & MPP\cite{5} & CWC (our) & Weeds\cite{weeds} & SugarBeet2017\cite{4} & CAW\cite{CAW} & RI\cite{6}\\
    \hline
     YOLO-DP& 11.7 & \textbf{1 (0.89)} & \textbf{1 (0.42)} & 2 (92.9\%) & 2 (76.5\%) & \textbf{1 (73.8\%)} & 2 (89.4\%) & \textbf{1 (82.3\%)}\\
     YOLOv3& 155.3 & 5 (0.88) & 5 (0.38) & 6 (86.6\%) & \textbf{1 (83.0\%)} & 6 (46.2\%) & 6 (73.1\%) & 6 (74.0\%)\\
     YOLOv4s& 20.8 & 3 (0.88) & 2 (0.42) & 5 (87.3\%) & 3 (75.6\%) & 3 (60.3\%) & 3 (86.5\%) & 2 (81.7\%)\\
     YOLOv5s& 16.0 & 2 (0.89) & 3 (0.38) & 4 (88.6\%) & 4 (75.3\%) & 4 (58.3\%) & 5 (82.5\%) & 3 (81.5\%)\\
     YOLOv7& 105.1 & 6 (0.85) & 6 (0.24) & \textbf{1 (93.1\%)} & 5 (74.4\%) & 5 (48.1\%) & 4 (83.3\%) & 4 (80.3\%)\\
     YOLOv8s& 28.6 & 4 (0.88) & 4 (0.38) & 3 (92.0\%) & 6 (70.4\%) & 2 (65.0\%) & \textbf{1 (90.1\%)} & 5 (79.3\%)\\
    \hline
  \end{tabular}}
\end{table}

\textbf{Comparative Experiment.}
We perform a comparative analysis of single-stage detectors including the full YOLO family, SSD, EfficientDet, and RetinaNet, the two-stage detector Fast-RCNN, anchor-free based CenterNet, and WeedNet-R, a specialized model for weed detection.
Among them, some models use pre-trained weights.
As some models do not provide the computation results for certain metricss, these are denoted by ``-'' in the table.

In \cref{tab:6}, we present the comparative evaluations of the proposed YOLO-DP baseline model alongside other models across the three datasets.
The YOLO-DP model does not need to load pre-trained weights, and the detection speed is significantly ahead of other models.
Under this premise, YOLO-DP demonstrates the most outstanding overall performance on the PDT dataset, with its F1 composite metrics surpassing all other models.
This shows that the YOLO-DP model can well adapt to the detection scenario of UAV with large-scale, high-precision, dense and small targets.
On the CWC dataset, its P metrics is at the first-class level, which is 4.3\% higher than the YOLOv5s baseline model.
This demonstrates that the YOLO-DP model possesses an effective capability for extracting fine-grained texture information and is capable of fulfilling the classification task requirements within this domain.
On the authoritative dataset SugarBeet2017, while the overall performance does not match that of models with loaded and trained weights, the R metrics of YOLO-DP is comparable to the pre-trained model's performance. 
This substantiates that the YOLO-DP model possesses precise detection capabilities and can accurately pinpoint the detection targets.

The comparative experimental results show that the YOLO-DP model is suitable for the application of plant protection UAV in precision agriculture detection fields such as weeds, pests and diseases, and crops.
\begin{table}[tb]
  \caption{Comparative experiment.}
  \begin{center}
  \label{tab:6}
  \resizebox{0.75\linewidth}{!}{
  \begin{tabular}{@{}ccccccccccc@{}}
    \toprule
     Datasets & Approach & P & R & mAP@.5 & mAP@.5:.95 & F1 & Gflops & Parameters & FPS & Pre-training\\
    \midrule
     \multirow{13}{*}{\makecell{PDT \\dataset \\(LL)}} & SSD\cite{SSD} & 84.5\% & 87.7\% & 85.1\% & - & 0.86 & 273.6 & 23.7M & 37 & \checkmark\\
     & EfficientDet\cite{eff} & 92.6\% & 73.4\% & 72.3\% & - & 0.82 & \textbf{11.5} & 6.7M & 12 & \checkmark\\
     & RetinaNet\cite{retina} & 93.3\% & 65.3\% & 64.2\% & - & 0.79 & 109.7 & 32.7M & 32 & \checkmark\\
     & CenterNet\cite{center} & \textbf{95.2\%} & 67.4\% & 66.5\% & - & 0.79 & 109.7 & 32.7M & 32 & \checkmark\\
     & Faster-RCNN\cite{faster} & 57.8\% & 70.5\% & 61.7\% & - & 0.64 & 401.7 & 136.7M & 13 & \checkmark\\
     & YOLOv3\cite{v3} & 88.5\% & 88.1\% & 93.4\% & 65.7\% & 0.88 & 155.3 & 61.5M & 41 & -\\
     & YOLOv4s\cite{v4} & 88.8\% & 88.2\% & \textbf{94.7\%} & 66.1\% & 0.88 & 20.8 & 9.1M & 51 & -\\
     & YOLOv5s\_7.0\cite{YOLOv5} & 88.9\% & \textbf{88.5\%} & 94.2\% & 67.0\% & \textbf{0.89} & 16.0 & 7.0M & 93 & -\\
     & YOLOv6s\cite{v6} & - & - & 91.4\% & 63.2\% & - & 44.1 & 17.2M & 43 & -\\
     & YOLOv7\cite{v7} & 87.4\% & 82.6\% & 90.1\% & 55.5\% & 0.85 & 105.1 & 37.2M & 32 & -\\
     & YOLOv8s\cite{v8} & 88.7\% & 87.5\% & 94.0\% & \textbf{67.9\%} & 0.88 & 28.6 & 11.1M & 60 & -\\
     & WeedNet-R\cite{4} & 87.7\% & 48.1\% & 70.4\% & - & 0.62 & 19.0 & 25.6M & 0.5 & -\\
     & YOLO-DP (our) & 90.2\% & 88.0\% & 94.5\% & 67.5\% & \textbf{0.89} & 11.7 & \textbf{5.2M} & \textbf{109} & -\\
     \hline
     \multirow{12}{*}{\makecell{CWC \\dataset}} & SSD\cite{SSD} & \textbf{97.7\%} & 77.6\% & 85.7\% & - & 0.91 & 426.9 & 23.7M & 29 & \checkmark\\
     & EfficientDet\cite{eff} & 97.2\% & \textbf{98.6\%} & \textbf{98.6\%} & - & 0.90 & \textbf{11.5} & 6.7M & 13 & \checkmark\\
     & RetinaNet\cite{retina} & 95.1\% & 98.3\% & 98.0\% & - & \textbf{0.97} & 261.3 & 36.4M & 24 & \checkmark\\
     & CenterNet\cite{center} & 96.6\% & 73.8\% & 73.3\% & - & 0.80 & 171.4 & 32.7M & 27 & \checkmark\\
     & YOLOv3\cite{v3} & 86.8\% & 89.4\% & 93.2\% & 82.3\% & 0.88 & 154.7 & 61.5M & 30 & -\\
     & YOLOv4s\cite{v4} & 87.3\% & 87.9\% & 91.9\% & 81.5\% & 0.88 & 20.8 & 9.1M & 43 & -\\
     & YOLOv5s\_7.0\cite{YOLOv5} & 88.6\% & 88.7\% & 93.0\% & 81.2\% & 0.89 & 16.0 & 7.0M & 65 & -\\
     & YOLOv6s\cite{v6} & - & - & 92.7\% & 84.3\% & - & 68.9 & 17.2M & 31 & -\\
     & YOLOv7\cite{v7} & 93.1\% & 76.4\% & 88.1\% & 75.6\% & 0.84 & 105.1 & 37.2M & 21 & -\\
     & YOLOv8s\cite{v8} & 92.0\% & 89.1\% & 94.0\% & \textbf{86.2\%} & 0.91 & 28.6 & 11.1M & 38 & -\\
     & WeedNet-R\cite{4} & 86.1\% & 51.8\% & 71.6\% & - & 0.65 & 19.0 & 25.6M & 0.5 & -\\
     & YOLO-DP (our) & 92.9\% & 87.5\% & 91.8\% & 81.0\% & 0.90 & 11.7 & \textbf{5.2M} & \textbf{72} & -\\
     \hline
     \multirow{12}{*}{\makecell{Sugar-\\Beet2017}} & SSD\cite{SSD} & 85.0\% & 83.6\% & 79.3\% & - & 0.85 & 1120 & 23.7M & 19 & \checkmark\\
     & EfficientDet\cite{eff} & 93.3\% & 79.8\% & 77.8\% & - & \textbf{0.86} & \textbf{11.5} & 6.7M & 16 & \checkmark\\
     & RetinaNet\cite{retina} & 91.7\% & 78.8\% & 76.6\% & - & 0.84 & 256.4 & 36.3M & 23 & \checkmark\\
     & CenterNet\cite{center} & \textbf{97.9\%} & 51.2\% & 51.0\% & - & 0.62 & 117.4 & 32.7M & 41 & \checkmark\\
     & Faster-RCNN\cite{faster} & 63.6\% & \textbf{87.4\%} & \textbf{80.0\%} & - & 0.73 & 546.9 & 136.7M & 25 & \checkmark\\
     & YOLOv3\cite{v3} & 34.8\% & 46.2\% & 39.4\% & 25.6\% & 0.40 & 155.3 & 61.5M & 28 & -\\
     & YOLOv4s\cite{v4} & 28.1\% & 60.3\% & 41.1\% & 26.4\% & 0.38 & 20.8 & 9.1M & 28 & -\\
     & YOLOv5s\_7.0\cite{YOLOv5} & 25.0\% & 58.3\% & 40.6\% & \textbf{26.7\%} & 0.35 & 16.0 & 7.0M & 50 & -\\
     & YOLOv6s\cite{v6} & - & - & 24.6\% & 15.0\% & - & 185.2 & 17.2M & 49 & -\\
     & YOLOv7\cite{v7} & 34.2\% & 48.1\% & 38.6\% & 24.9\% & 0.40 & 105.1 & 37.2M & 18 & -\\
     & YOLOv8s\cite{v8} & 23.9\% & 65.0\% & 39.1\% & 26.1\% & 0.35 & 28.6 & 11.1M & 33 & -\\
     & WeedNet-R\cite{4} & 90.1\% & 68.4\% & 84.8\% & - & 0.78 & 19.0 & 25.6M & 0.5 & -\\
     & YOLO-DP (our) & 23.1\% & 73.8\% & 38.3\% & 25.0\% & 0.35 & 11.7 & \textbf{5.2M} & \textbf{62} & -\\
    \bottomrule
  \end{tabular}}
  \end{center}
\end{table}

\textbf{Ablation Experiment.}
We chose the YOLOv5s that performed well on the PDT dataset as the benchmark. 
We use the popular attention mechanism to ablate the adaptive large-scale selection kernel to verify its detection performance on PDT dataset.
The attention modules are designed on BackBone's floors 3, 5, 7, and 9. Since the number of C3x and C3TR parameters is too large, we only replace them at the 9th layer in order to carry out the experiment.
“v5s'' means YOLOv5s\_7.0. “v5s\_our'' stands for the use of Adaptive Large Scale Selective Kernel in the YOLOv5s\_7.0.
It can be seen from the results in \cref{tab:7} that the performance of YOLOv5s is greatly improved after Large Scale Selective Kernel is used. 
The mAP@.5, mAP@.5:.95 and F1 are the best metrics. 
This indicates that the Large Scale Selective Kernel used in this paper is suitable for dense object detection tasks.
\begin{table}[tb]
  \caption{Ablation experiment.}
  \begin{center}
  \label{tab:7}
  \resizebox{0.6\linewidth}{!}{
  \begin{tabular}{@{}ccccccccc@{}}
    \toprule
    Datasets & Approach & P & R & mAP@.5 & mAP@.5:.95 & F1 & Gflops & Parameters\\
    \midrule
     \multirow{13}{*}{\makecell{PDT \\dataset \\(LL)}} & v5s\_C1 & 88.2\% & 88.5\% & 93.9\% & 67.1\% & 0.88 & 25.3 & 10.0M\\
      & v5s\_C2 & 88.8\% & 88.4\% & 94.1\% & 67.0\% & 0.88 & 17.2 & 7.3M\\
      & v5s\_C2f & 88.6\% & 88.5\% & 93.8\% & 67.1\% & 0.88 & 17.4 & 7.5M\\
      & v5s\_C3 & 88.9\% & 88.5\% & 94.2\% & 67.0\% & \textbf{0.89} & 16.0 & 7.0M\\
      & v5s\_C3x & 88.7\% & 81.5\% & 87.4\% & 63.2\% & 0.85 & 14.5 & 6.5M\\
      & v5s\_C3TR & 88.3\% & \textbf{89.0\%} & 94.1\% & 67.1\% & \textbf{0.89} & 15.7 & 7.0M\\
      & v5s\_C3Ghost & 88.9\% & 88.2\% & 94.2\% & 66.7\% & 0.88 & 12.5 & 5.9M\\
      & v5s\_SE & 88.8\% & 88.3\% & \textbf{94.4\%} & 66.6\% & 0.88 & 10.6 & \textbf{5.1M}\\
      & v5s\_CBAM & \textbf{89.8\%} & 87.5\% & \textbf{94.4\%} & 66.5\% & \textbf{0.89} & 10.9 & 5.6M\\
      & v5s\_GAM & 89.2\% & 87.7\% & 94.0\% & 67.1\% & 0.88 & 16.4 & 7.5M\\
      & v5s\_ECA & 89.7\% & 87.0\% & 94.3\% & 66.2\% & 0.88 & \textbf{10.5} & \textbf{5.1M}\\
      & v5s\_our & 89.1\% & 88.5\% & \textbf{94.4\%} & \textbf{67.2\%} & \textbf{0.89} & 12.2 & 6.1M\\
    \bottomrule
  \end{tabular}}
  \end{center}
\end{table}

\subsection{Visualization Research}
\begin{figure}[tb]
  \centering
  \includegraphics[width=0.95\linewidth]{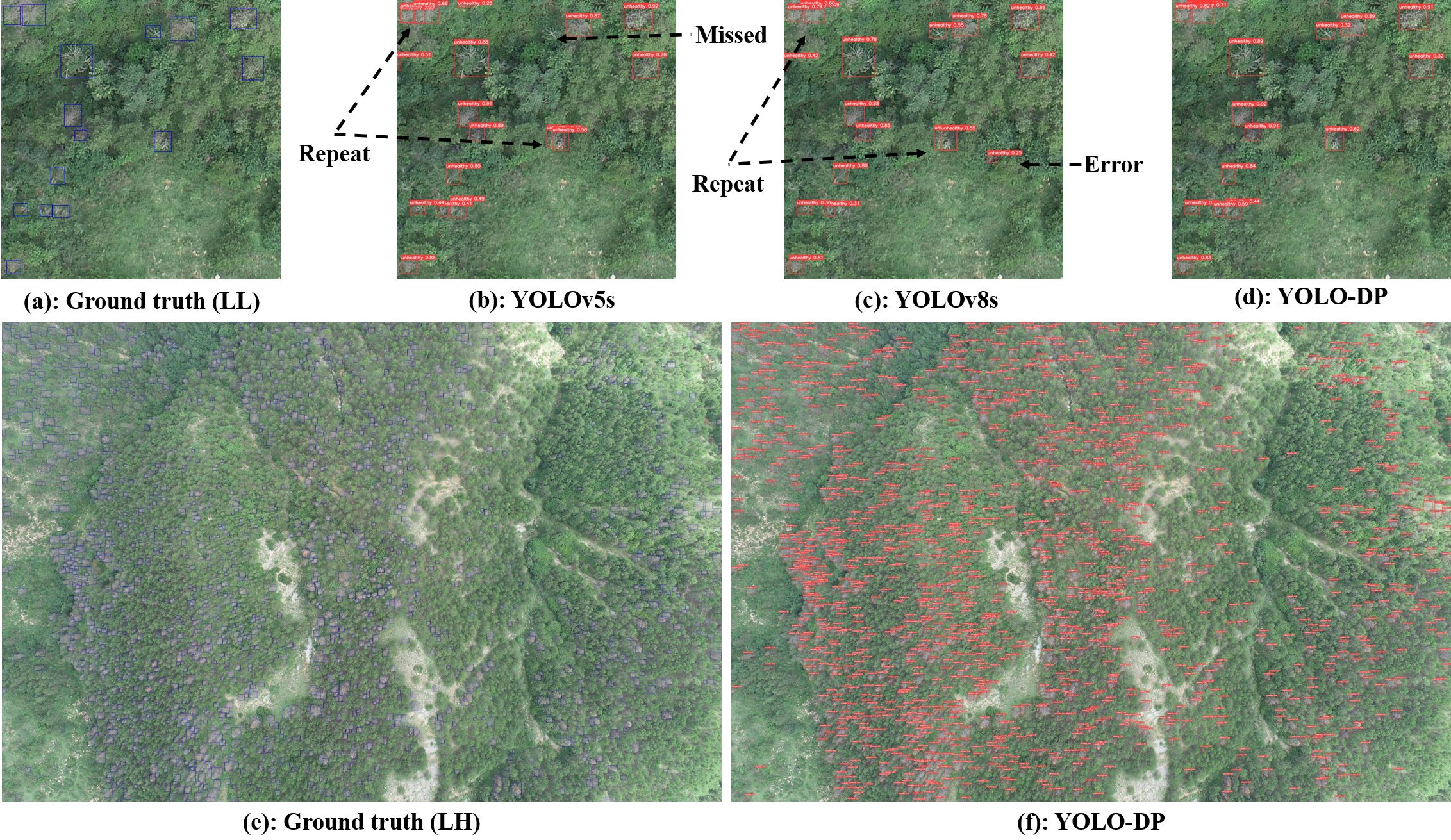}
  \caption{Visualization of PDT dataset detection.}
  \label{fig:ksh_LL_LH}
\end{figure}

In \cref{fig:ksh_LL_LH},  we show the detection visualization results on the PDT dataset.
We have selected the YOLOv5s and YOLOv8s models, which exhibited superior overall performance in the comparative experiment, to demonstrate and contrast with the YOLO-DP model. 
(a) and (e) illustrate the Ground Truth for the LL and LH versions, respectively. 
Observations indicate that the "Human-in-the-loop'' data annotation approach is practical for weed and pest detection data annotation.
The detection results in (b) and (c) suggest that the LL version of PDT dataset is compatible with widely used detection models, confirming its utility. 
When compared to (d), it is evident that YOLO-DP delivers outstanding detection performance on LL, with no instances of missing, false, or duplicate detections. 
Examining the LH version's detection results in (f), we find no missed detections, and the critical issue of current detectors' inability to distinguish dead pine trees from the ground has been overcome. 
This leads us to conclude that the YOLO-DP model is well-suited for large-scale, high-precision, small-target UAV detection tasks. 
Moreover, the detection strategy depicted in \cref{fig:jianceliucheng} of this paper has been validated as effective.

\begin{figure}[tb]
  \centering
  \includegraphics[width=1\linewidth]{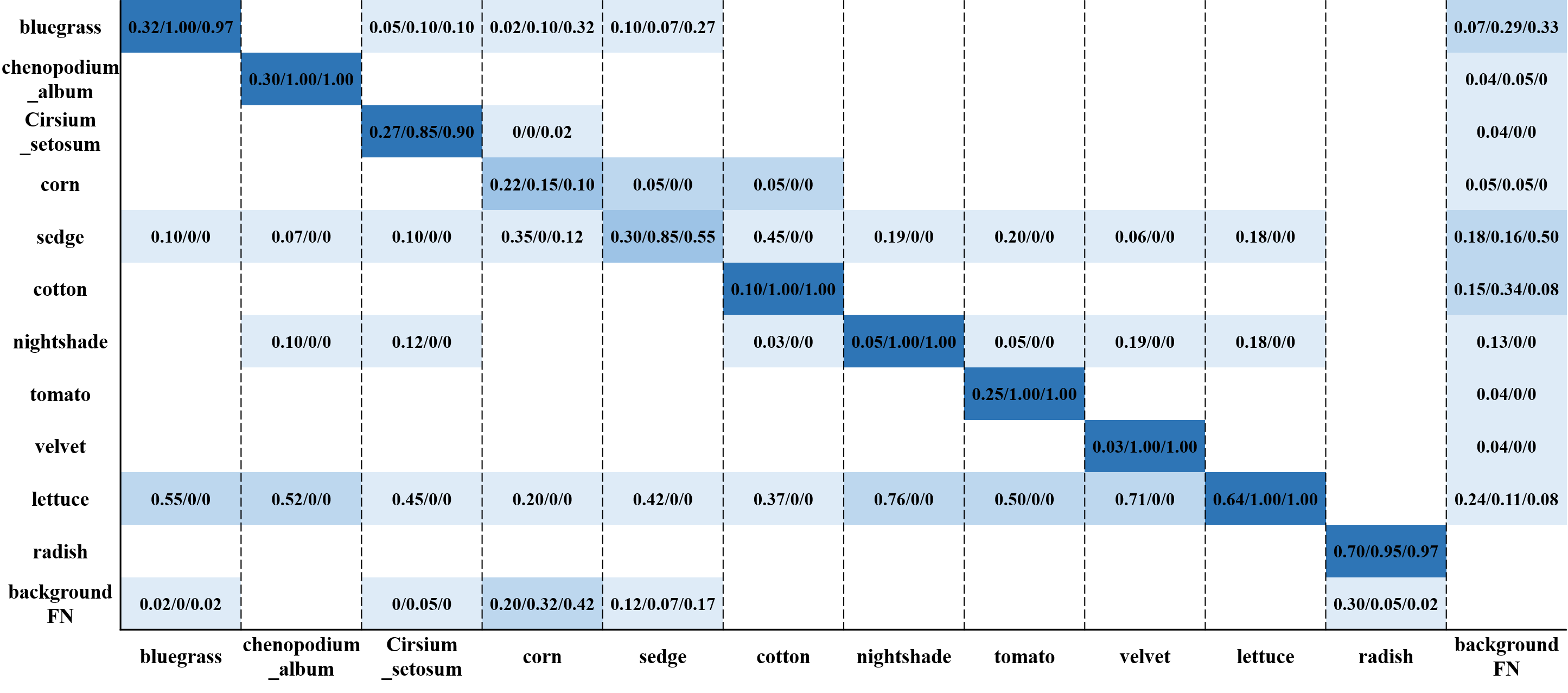}
  \caption{Confusion matrix.
  The rows represent the true class, the columns represent the predicted class, and the confidence value is [0,1]. The format of the matrix is: YOLOv7/YOLOv8s/YOLO-DP.}
  \label{fig:Labels statistics}
\end{figure}

In \cref{fig:Labels statistics}, we presents a comparison between YOLO-DP and the Confusion Matrix of the high-performing classification models YOLOv7 and YOLOv8s on the CWC dataset.
It is evident that YOLO-DP's classification performance surpasses that of YOLOv7.  
Furthermore, YOLO-DP's classification capability is comparable to YOLOv8s, especially when its overall performance exceeds that of YOLOv8s.

\section{Conclusion}
In this study, we introduce the PDT dataset to address the challenging task of UAV pest detection, with the goal of advancing research in UAV detection tasks within the realm of precision agriculture. We also present the CWC dataset for agricultural weed and crop classification, which compensates for the field's deficiency in training model classification capabilities. To tackle the scarcity of baseline model research in this domain, we propose the YOLO-DP model for dense, small-target UAV pest detection and validate its efficacy across three datasets. It is noteworthy that the PDT dataset features a dual-resolution version, and the CWC dataset boasts 11 detailed, texture-similar plant classes. Moreover, we provide a comprehensive evaluation of both datasets, objectively assessing their value. This work has its limitations: the 3D point cloud version of the PDT dataset is not discussed, and there is a size inconsistency issue within the CWC dataset. We plan to address these issues in future endeavors. Lastly, to ensure the continuity of this research, we make the datasets and associated code designed in this paper available on our website.

\section*{Acknowledgments}
This work was supported by Key R\&D Program of Shandong Province, China (2023CXGC010112), the Taishan Scholars Program (NO. tsqn202103097, NO. tscy20221110).

%
%
\bibliographystyle{splncs04}

\newpage

\appendix

\section{Data Acquisition}
\begin{figure}[tb]
  \centering
  \includegraphics[width=1\linewidth]{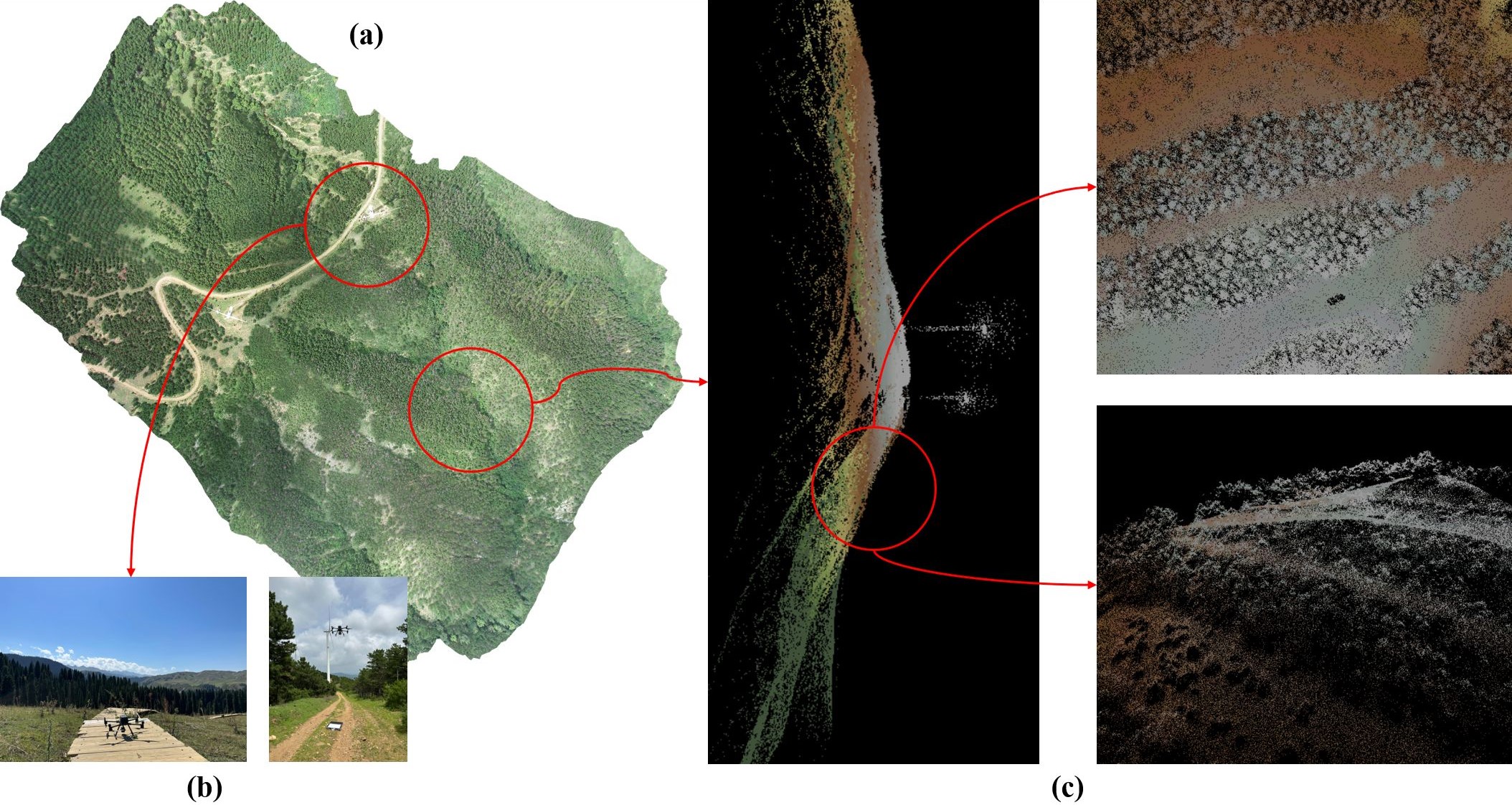}
  \caption{Visualization of data acquisition. The source landform, shooting process and 3D point cloud data of PDT dataset are presented.}
  \label{fig:caiji}
\end{figure}

\cref{fig:caiji} shows the details of the data acquisition for this work. (a) Orthophoto images from drones with ultra-high precision. (b) demonstrates our data acquisition process using DJI-ChanSi L2 equipment. (c) consists of two parts, with the left half showing an overview of the 3D point cloud data and modeling the entire mountain range. The right half shows the details of 3D point cloud data, and the characteristics of high resolution of its data can be observed.

\section{Crop and Weed Detection Data and Weeds Datasets}
In order to further verify the performance of YOLO-DP, the experiment is more convincing. We conducted experiments on two other publicly available datasets.

\begin{figure}[tb]
  \centering
  \includegraphics[width=1\linewidth]{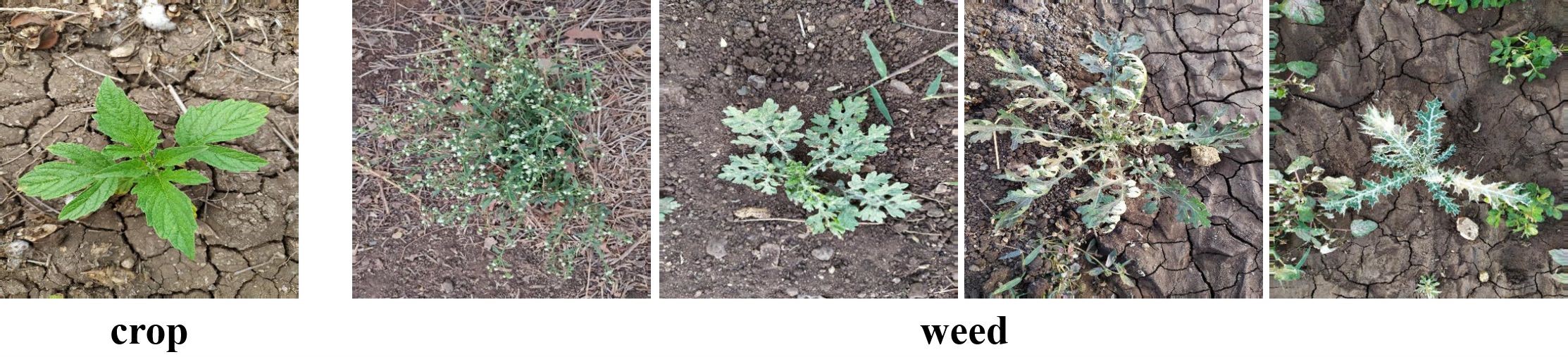}
  \caption{Visualization of Crop and Weed Detection Data. The data set is characterized by abundant detailed texture information.}
  \label{fig:shujvji1}
\end{figure}

 \textbf{Crop and Weed Detection Data:} The dataset contains 1,300 images of sesame crops and different types of weeds (\cref{fig:shujvji1}). Each image has a label, and the image label is in the YOLO format. The dataset contains two categories, crop and weed (weed is made up of a variety of plants). The image size is 512$\times$512. This dataset is characterized by rich texture information, which can test the classification ability of the model.

\begin{figure}[tb]
  \centering
  \includegraphics[width=1\linewidth]{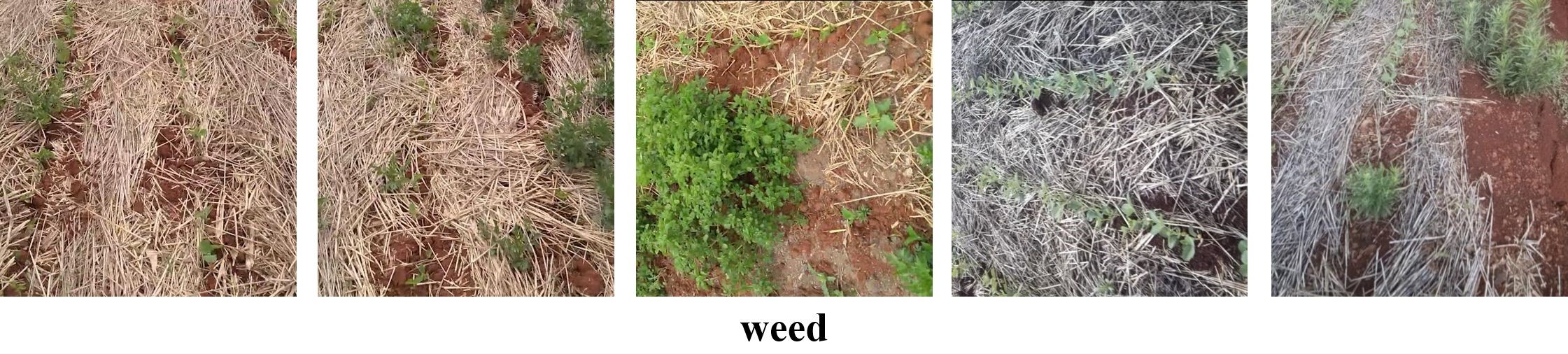}
  \caption{Visualization of Weeds. The data set is characterized by a realistic detection environment.}
  \label{fig:shujvji2}
\end{figure}

 \textbf{Weeds:} This public dataset uses a database of photos of weeds detected in soybean plantations (\cref{fig:shujvji2}). The original images in the database used were produced and provided by Eirene Solutions. The database consists of two parts: 92 photos collected by photographers at different points on soybean plantations and 220 video frames of photographers walking on soybean and corn plantations. The dataset contains one category: weed, and the image size is 480$\times$480. The characteristic of this data set is to simulate the real environment, which can test the detection ability of the model.

\section{Experiment Indicators}
In this section, we introduce the experimental indicators used in this paper.

\textbf{P(Precision):} The accuracy rate is the ratio of the number of correct tests to the number of positive tests.  Equation~(\ref{e14}) is as follows:
\begin{equation}
\label{e14}
{ P }=\frac{T P}{T P+F P}
\end{equation}

 \textbf{R(Recall):} Recall is the ratio of the number of correct detections to the number of actual positive detections.  Equation~(\ref{e15}) is as follows:
\begin{equation}
\label{e15}
{ R }=\frac{T P}{T P+F N}
\end{equation}
where $TP$ indicates the number of actuals and detections that are positive;  $FP$ indicates the number of detections that are positive but negative;  $FN$ indicates the number of detections that are negative but positive. 

\textbf{mAP(Mean Average Precision):} mAP is the mean value of the average precision of the different categories.  Equation~(\ref{e17}) is as follows:
\begin{equation}
\label{e17}
m A P=\frac{\sum_{1}^{N} \int_{0}^{1} p(r) d r}{N} 
\end{equation}
where $N$ represents the total number of categories of detection objects, $p$ represents the precision of each category, and $r$ represents the recall of each category.

\textbf{F1:} F1 is a measure of the accuracy of a model that takes into account both the accuracy and recall of a classification model.  Equation~(\ref{e16}) is as follows:
\begin{equation}
\label{e16}
F 1 =\frac{2 \times { P } \times { R }}{ { P }+ { R }}
\end{equation}

\textbf{GFLOPs (Giga Floating-point Operations Per Second):} The number of floating-point operations per second can be used to measure the complexity of a model. Smaller GFLOPs indicate that the model requires less computation and runs faster.

\textbf{Parameters:} Parameters refers to the total number of parameters to be trained in model training. It is used to measure the size of the model. The unit is usually M. M refers to millions and is the unit of count.

\textbf{FPS:} The number of frames transmitted per second, how many frames (how many pictures) the network can process (detect) per second, that is, the number of pictures that can be processed per second or the time it takes to process an image to evaluate the detection speed, the shorter the time, the faster the speed.

\section{Hyperparameter Selection}
The choice of hyperparameters is not arbitrary. We have already experimented with YOLO-DP.
\cref{tab:2.0-a} is the main result of hyperparameter experiments in the PDT dataset (LL). $lr_0$ is the initial learning rate, $lr_f$ is the cycle learning rate, $bs$ is the batch-size, $ws$ is the workers.
\begin{table}[tb]
    \vspace{-9pt}
  \caption{Hyperparameter experiment on PDT dataset (LL).}
  \label{tab:2.0-a}
  \centering
  \resizebox{0.7\linewidth}{!}{
  \begin{tabular}{@{}ccccccc@{}}
    \hline
    \multicolumn{2}{c}{Hyperparameter} & P & R & mAP@.5 & mAP@.5:.95 & F1\\
    \hline
     \multirow{3}{*}{\textbf{$\boldsymbol{lr_0}$: 0.01+$\boldsymbol{lr_f}$: 0.01}}& $bs$: 8 + $ws$: 4& 90.1\% & 86.8\% & 94.3\% & 66.4\% & 0.88\\
     & \textbf{$\boldsymbol{bs}$: 16 + $\boldsymbol{ws}$: 8}& \textbf{90.2\%} & \textbf{88.0\%} & \textbf{94.5\%} & \textbf{67.5\%} & \textbf{0.89}\\
     & $bs$: 32 + $ws$: 16& 89.5\% & 87.1\% & 94.2\% & 66.4\% & 0.88\\
     \hline
     $lr_0$: 0.01+$lr_f$: 0.1& \multirow{2}{*}{$bs$: 16 + $ws$: 8}& 89.0\% & 87.0\% & 93.9\% & 65.5\% & 0.88\\
     $lr_0$: 0.01+$lr_f$: 0.001& & 90.0\% & 86.8\% & 94.1\% & 66.1\% & 0.88\\
    \hline
  \end{tabular}}
\end{table}

\section{Comparative Experiment}
To make the comparative experiment of this work more convincing, we performed additional comparisons on the crop and weed detection data public datasets.
Among these, SSD, EfficientDet, RetinaNet, and CenterNet utilize pre-trained weights. 
As some models do not provide the computation results for certain parameters, these are denoted by $"$-$"$ in the table.

From the data in \cref{tab:2-a}, we can observe that the comprehensive level of our YOLO-DP model has reached the best. Among them, mAP@.5 index reached the highest 93.3\%.
\begin{table}[tb]
  \caption{Comparative experiment of Crop and Weed Detection Data.}
  \begin{center}
  \label{tab:2-a}
  \resizebox{0.6\linewidth}{!}{
  \begin{tabular}{@{}cccccccc@{}}
    \toprule
    Approach & P & R & mAP@.5 & mAP@.5:.95 & F1 & Pre-training\\
    \midrule
     SSD & 74.6\% & 84.6\% & 74.2\% & - & 0.79 & \checkmark\\
     EfficientDet & 90.9\% & 76.4\% & 73.9\% & - & 0.83 & \checkmark\\
     RetinaNet & 87.0\% & 76.4\% & 73.9\% & - & 0.81 & \checkmark\\
     CenterNet & \textbf{100.0\%} & 28.7\% & 28.7\% & - & 0.41 & \checkmark\\
     Faster-RCNN & 39.5\% & \textbf{90.1\%} & 81.7\% & - & 0.55 & \checkmark\\
     \hline
     YOLOv3 & 92.1\% & 73.1\% & 85.2\% & 46.4\% & 0.78 & -\\
     YOLOv4s & 91.6\% & 86.5\% & 93.0\% & 57.6\% & \textbf{0.88} & -\\
     YOLOv5s\_7.0 & 88.0\% & 82.5\% & 91.2\% & 56.6\% & 0.85 & -\\
     YOLOv6s & - & - & 89.3\% & 55.3\% & - & -\\
     YOLOv7 & 93.8\% & 83.3\% & 90.4\% & 53.7\% & \textbf{0.88} & -\\
     YOLOv8s & 87.5\% & \textbf{90.1\%} & 90.9\% & \textbf{61.2\%} & 0.86 & -\\
     WeedNet-R & 87.5\% & 63.5\% & 77.6\% & - & 0.73 & -\\
     YOLO-DP (our) & 84.1\% & 89.4\% & \textbf{93.3\%} & 53.3\% & 0.87 & -\\
    \bottomrule
  \end{tabular}}
  \end{center}
\end{table}


\section{Ablation Experiment}
In this section, we add ablation experiments on the YOLO-DP model. We chose the  YOLOv8s models that performed well on the Weeds dataset as the benchmark. We designed to use a popular attention mechanism to ablate the Adaptive Large Scale Selective Kernel in this paper to demonstrate its detection performance on UAV large-scale dense small-target pest data. Notably, the attention modules are designed on BackBone's floors 3, 5, 7, and 9. Since the number of C3x and C3TR parameters is too large, we only replace them at the 9th layer in order to carry out the experiment.
\begin{table}[tb]
\caption{Ablation experiment of Weeds. "v8s" means YOLOv8s. v8s\_DP stands for the use of Adaptive Large Scale Selective Kernel in the YOLOv8s benchmark model.}
\begin{center}
\resizebox{0.6\linewidth}{!}{
\begin{tabular}{@{}cccccccc@{}}
\toprule
Approach & P  & R  & mAP@.5 & mAP@.5:.95  & F1 & Gflops & Parameters\\ 
\midrule
v8s\_C1   & 70\%   & 64.2\%   & 70.1\%   & 37.6\%   & 0.67  & 32.6  & 12.5M \\
v8s\_C2   & 77.5\%   & 64.2\%   & 72\%   & \textbf{39.6\%}   & 0.70  & 27.8  & 10.9M \\
v8s\_C2f   & 70.4\%   & 63.3\%   & 69\%   & 36.9\%   & 0.67  & 28.6  & 11.1M \\
v8s\_C3   & 71.7\%   & 64.9\%   & 72.5\%   & 38.2\%   & 0.69  & 25.3  & 10.0M \\
v8s\_C3x   & 74.5\%   & 65\%   & 70\%   & 38\%   & 0.69  & 24.0  & 9.6M \\
v8s\_C3TR   & 75.2\%   & 59.9\%   & 69.3\%   & 36.7\%   & 0.66  & 27.7  & 10.4M \\
v8s\_C3Ghost   & 64.4\%   & \textbf{68.2\%}   & 71.2\%   & 38.9\%   & 0.66  & 22.4  & 9.4M \\
v8s\_SE   & 75.3\%   & 66.8\%   & 72.4\%   & 38.8\%   & \textbf{0.71}  & \textbf{20.5}  & 8.3M \\
v8s\_CBAM   & \textbf{80.4\%}   & 61.3\%   & 72.9\%   & 38.9\%   & 0.69  & 20.8  & 8.7M \\
v8s\_GAM   & 69\%   & 65\%   & 68.5\%   & 37.9\%   & 0.67  & 28.4  & 11.0M \\
v8s\_ECA   & 76.3\%   & 64.6\%   & 71.2\%   & 38.4\%   & 0.70  & \textbf{20.5}  & \textbf{8.2M} \\
v8s\_DP  & 71.5\%   & 67.2\%   & 72.5\%   & 38.6\%   & 0.69  & 23.1  & 9.0M \\
YOLO-DP (our)  & 73.1\%   & 66.1\%   & \textbf{73.1\%}   & 38.1\%   & 0.69  & 22.5  & 8.6M \\
\bottomrule
\end{tabular}}
\end{center}
\label{tab:4-a}
\end{table}

\cref{tab:4-a} shows that the performance of YOLOv8s is not the best. However, compared with using the C2f attention mechanism, the performance of the YOLOv8s benchmark model is greatly improved. Among them, P is increased by 1\%, R by 4\%, mAP@.5 by 3.5\%, mAP@.5:.95 by 2\%, F1 by 0.02, and the model complexity is reduced.

\begin{figure}[p]
  \centering
  \includegraphics[width=1\linewidth]{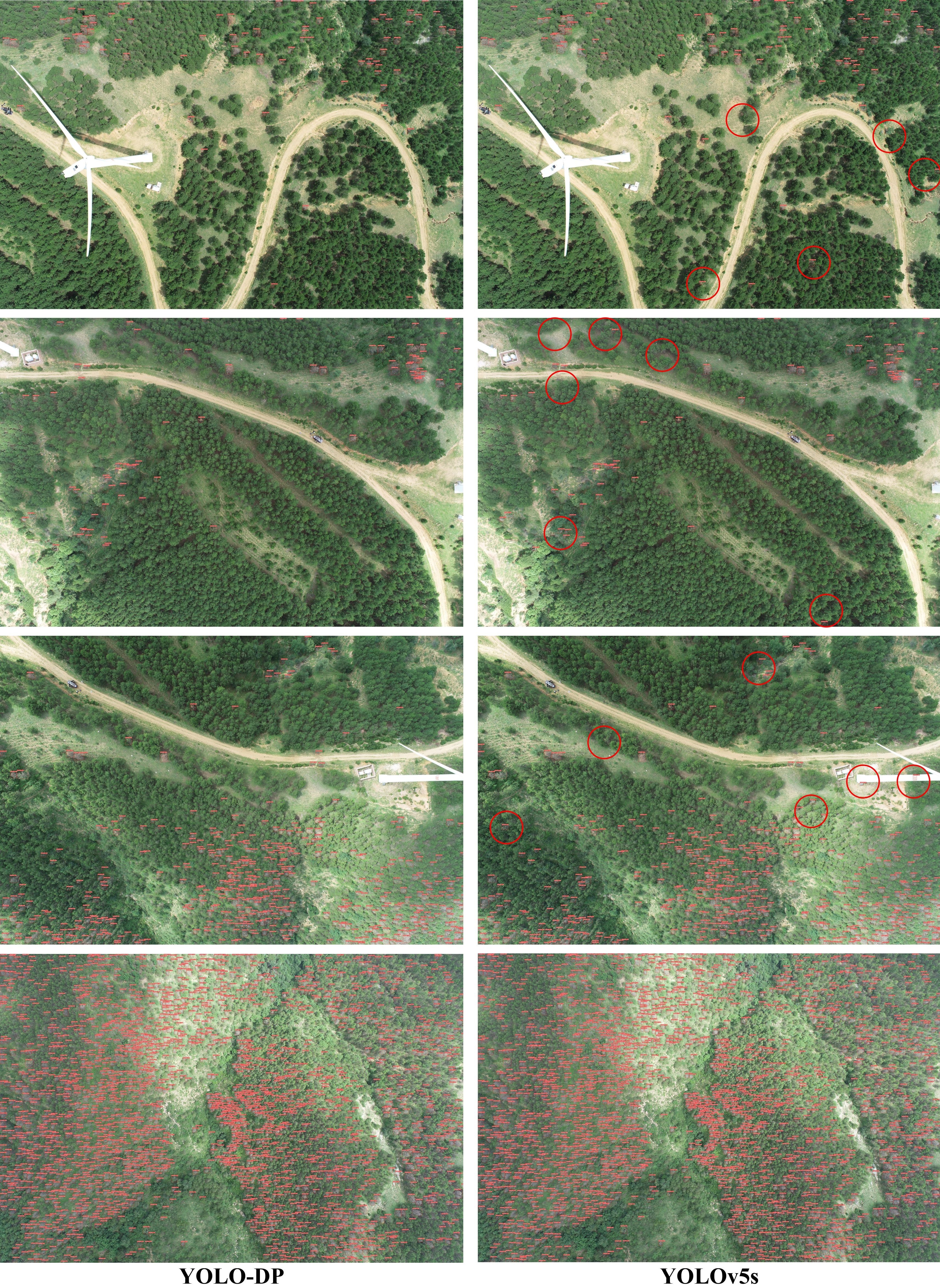}
  \caption{Visualization of detection result (LH). Three pairs of comparison figures, one pair of comparison figures of dense detection effect. Image size: 5472$\times$3648.}
  \label{fig:jiance}
\end{figure}

\section{Visualization Research}

\cref{fig:jiance} adds the comparison of test results of LH for the PDT dataset. We chose YOLOv5s with excellent detection performance for comparison. It can be observed that in the first three comparison graphs, the YOLO-DP model has well solved the problems of missing detection and wrong detection of the existing model. The fourth comparison diagram shows the detection effect of the YOLO-DP and YOLOv5s drones under the perspective of large-scale and dense small targets.

\begin{figure}[p]
  \centering
  \includegraphics[width=1\linewidth]{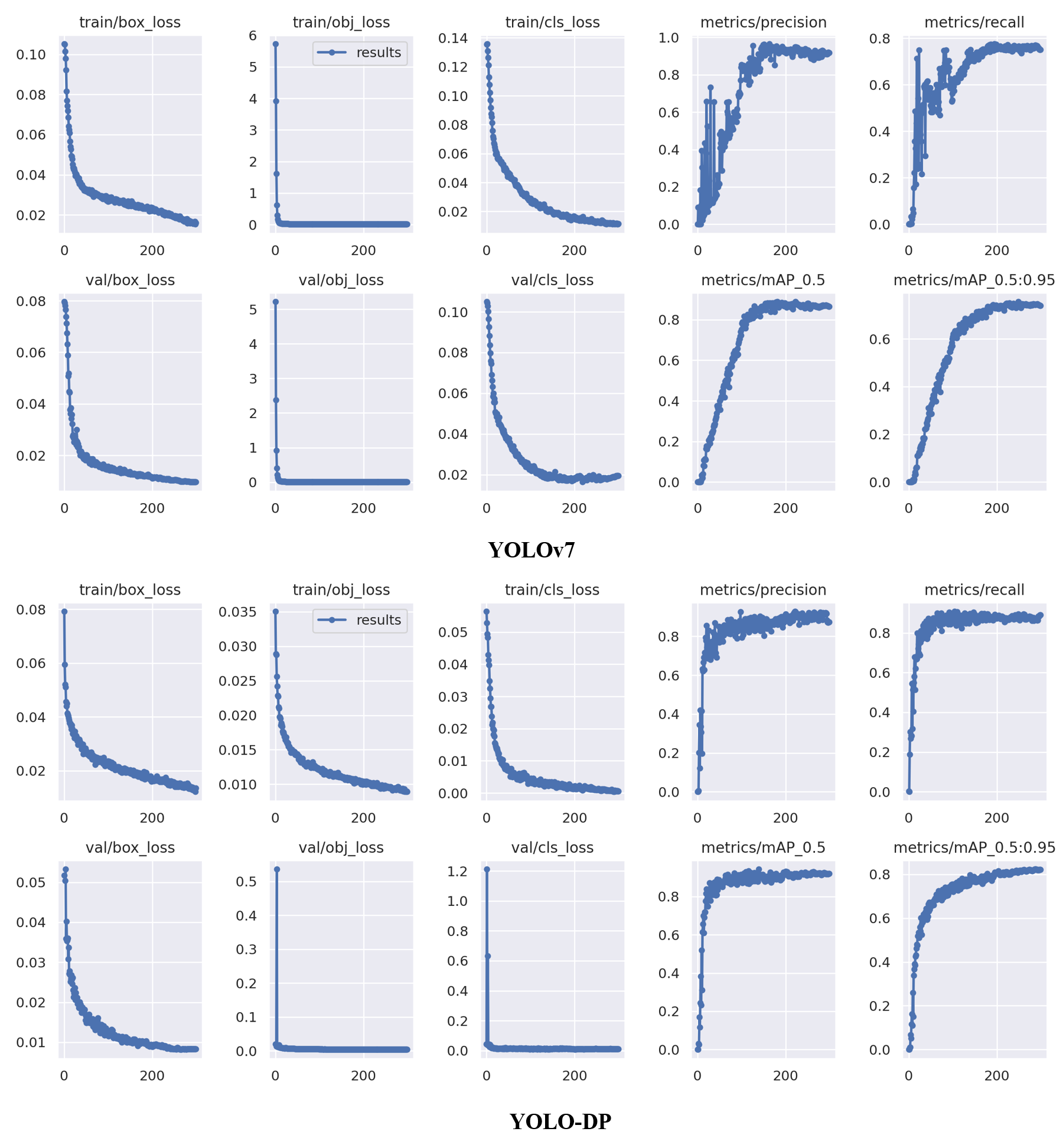}
  \caption{Visualization of CWC dataset training process. loss and other indicators in the training process.}
  \label{fig:loss}
\end{figure}

In the comparative experiment using CWC dataset to test YOLO-DP, we found that although the comprehensive performance of YOLOv7 was far behind that of YOLO-DP. However, its P parameter exceeds that of YOLO-DP model, and P parameter is an important index to measure the classification ability of the model. Therefore, we choose to compare the training loss and other indicators of the YOLOv7 model. It can be observed in \cref{fig:loss} that val/cls\_loss, train/cls\_loss, P, R, mAP@.5 and mAP@.5:.95 indicators of YOLO-DP model are superior to those of YOLOv7 model. This shows that the classification capability of YOLO-DP model can meet the actual demand.

\end{document}